%% file: main.tex
\documentclass{article}

\usepackage[preprint]{neurips_2026}

\makeatletter
\renewcommand{\@noticestring}{%
  \scriptsize
  $^{*}$Equal contribution\quad
  $^{\bigstar}$Project Lead\quad
  $^{\text{\Letter}}$Corresponding author%
}
\makeatother

\usepackage[utf8]{inputenc}
\usepackage[T1]{fontenc}
\usepackage{hyperref}
\usepackage{url}
\usepackage{booktabs}
\usepackage{amsfonts}
\usepackage{nicefrac}
\usepackage{microtype}
\usepackage{xcolor}
\usepackage{marvosym}
\usepackage{multirow}
\usepackage{booktabs}

\usepackage{graphicx}
\usepackage{amsmath,amssymb,mathtools}
\usepackage{enumitem}
\usepackage{xspace}
\usepackage{placeins}

\newcommand{\backbone}{\textsc{MindVLA-U1}\xspace}
\newcommand{\method}{\textsc{DIAL}\xspace}

\newcommand{\rfs}{\textsc{RFS}\xspace}
\newcommand{\grpo}{\textsc{GRPO}\xspace}
\newcommand{\sft}{\textsc{SFT}\xspace}

\newcommand{\vla}{\textsc{VLA}\xspace}

\title{Driving Intents Amplify Planning-Oriented Reinforcement Learning}

\author{%
  \parbox{\textwidth}{\centering
    Hengtong Lu$^{1,2*\text{\Letter}}$\quad
    Victor Shea-Jay Huang$^{1,3*}$\quad
    Chengmin Yang$^{1}$\quad
    Pengfei Jing$^{1,2}$\\
    Jifeng Dai$^{2}$\quad
    Yan Xie$^{1}$\quad
    Benjin Zhu$^{1,2*\text{\Letter}\bigstar}$\\[0.3em]
    {\mdseries\small $^{1}$Li Auto\quad
    $^{2}$Tsinghua University
    $^{3}$CUHK MMLab\quad\\[0.2em]
    \ttfamily
    luhengtong@lixiang.com,\quad
    zhubenjin@lixiang.com\\[0.2em]
    \normalfont\small
    Project page:\, \url{https://mind-omni.github.io/}}
  }
}

\begin{document}

\maketitle

\input{sections/00_abstract}
\input{sections/01_introduction}

\input{sections/02_related_work}
\input{sections/03_method}
\input{sections/04_experiments}
\input{sections/05_conclusion}
\bibliographystyle{plainnat}
\bibliography{references}

\newpage
\appendix
\input{sections/A_related_work}

\newpage
\input{sections/B_training_curves}

\end{document}

%% file: sections/00_abstract.tex
\begin{abstract}

Continuous-action policies trained on a single demonstrated trajectory per scene suffer from \emph{mode collapse}: samples cluster around the demonstrated maneuver and the policy cannot represent semantically distinct alternatives.
Under preference-based evaluation, this caps best-of-$N$ performance -- even oracle selection cannot recover what the sampling distribution does not contain.
We introduce DIAL, a two-stage \textbf{D}riving-\textbf{I}ntent-\textbf{A}mplified reinforcement \textbf{L}earning framework for preference-aligned continuous-action driving policies.
In the first stage, DIAL conditions the flow-matching action head on a discrete intent label with classifier-free guidance (CFG), which expands the sampling distribution along distinct maneuver modes and breaks single-demonstration mode collapse.
In the second stage, DIAL carries this expanded distribution into preference RL through \emph{multi-intent Group Relative Policy Optimization (GRPO)}, which spans all intent classes within every preference group and prevents fine-tuning from re-collapsing around the currently preferred mode.
Instantiated for end-to-end driving with eight rule-derived intents and evaluated on WOD-E2E: competitive Vision-to-Action (VA) and Vision-Language-Action (VLA) Supervised Finetuning (SFT) baselines plateau below the human-driven demonstration at best-of-$128$, with the strongest prior (RAP) capping at ater Feedback Score (RFS) $8.5$ even with best-of-$64$; intent-CFG sampling lifts this ceiling to RFS $9.14$ at best-of-$128$, surpassing both the prior best (RAP, $8.5$) and the human-driven demonstration ($8.13$) for the first time; and multi-intent GRPO improves held-out RFS from $7.681$ to $8.211$, while every single-intent baseline peaks lower and degrades by training end.
These results suggest that the bottleneck of preference RL on continuous-action policies trained from demonstrations is not only how to update the policy, but how to expand and preserve the sampling distribution being optimized.

\end{abstract}

%% file: sections/01_introduction.tex
\section{Introduction}
\label{sec:introduction}

Demonstrations record what happened, not what should have happened.
A logged trajectory is one physically feasible future realized in a particular scene.
Before the decision was made, several futures may still have been admissible: the vehicle may brake earlier or later, keep its lane or prepare a merge, proceed assertively or yield conservatively.
The demonstration certifies one executed behavior, not that this behavior is uniquely safe or most preferred by human evaluators.
That distinction is easy to miss when continuous-action policies are trained as geometric imitation, but it becomes central once policies are evaluated by preference alignment rather than by distance to a single recorded log.

\begin{figure*}[t]
  \centering
  \includegraphics[width=\linewidth]{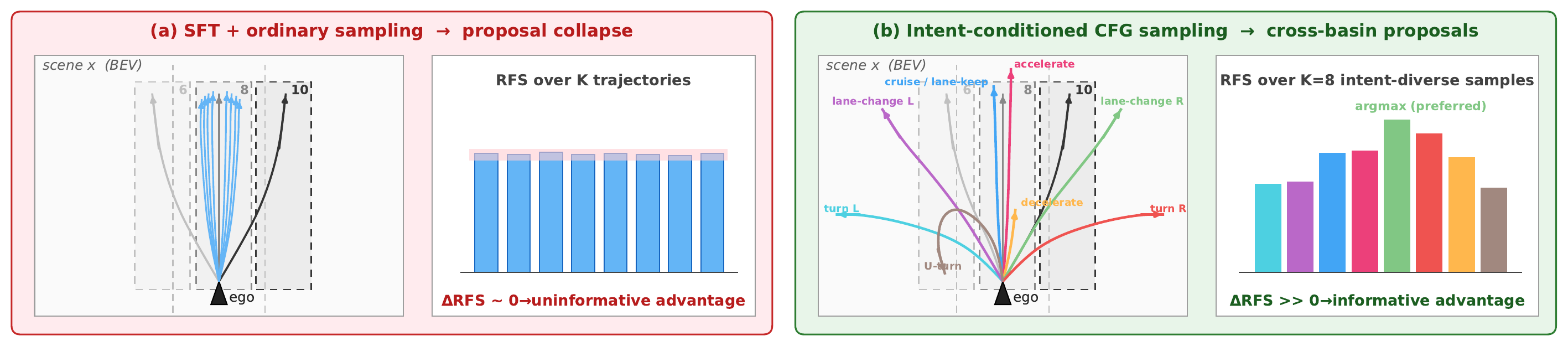}
  \caption{
    \textbf{Driving intents amplify planning-oriented RL by exposing within-scene preference contrast.}
    \textbf{(a)} Under \sft + ordinary sampling, $K$ rollouts collapse into one maneuver basin and their \rfs scores are nearly identical ($\Delta\rfs \approx 0$), so the group-relative advantage is uninformative.
    \textbf{(b)} Under intent-conditioned CFG sampling, $K{=}8$ rollouts (one per driving intent) spread across distinct basins and their \rfs scores spread widely ($\Delta\rfs \gg 0$), exposing the preference contrast that planning-oriented RL amplifies.
  }
  \label{fig:teaser}
\end{figure*}

Across robotic manipulation, navigation, and end-to-end driving \citep{zitkovich2023rt,kim2024openvla,zhou2025autovla,black2024pi_0,chi2025diffusion}, continuous-action policies are typically trained by imitation or supervised fine-tuning on a single demonstrated trajectory per scene.
With expressive flow-matching or diffusion action heads, this produces smooth local control near the demonstration but does little to expose alternative maneuvers: sampled trajectories cluster around the logged path, and the policy collapses to a single behavioral mode.
Multimodal trajectory representations have long addressed this for prediction \citep{chai2019multipath,phan2020covernet,salzmann2020trajectron++,shi2022motion}, but for the ego policy of an action planner, single-demonstration supervision remains the default.

Mode collapse is hard to see when policies are scored by distance to the demonstration, because the metric rewards exactly the behavior collapse concentrates on.
Preference-based evaluation makes it visible.
The WOD-E2E RFS benchmark \citep{xu2025wod} scores predicted trajectories against multiple human-annotated alternatives, so the logged GT is one scored candidate rather than an oracle target, and best-of-$K$ RFS over a policy's samples quantifies the rater-preferred quality reachable under oracle selection.
Across four competitive SFT VA and VLA baselines (Figure~\ref{fig:bon_intent_strategies}), best-of-$K$ RFS saturates \emph{below} the logged G round Truth (GT) (RFS $8.13$) even at $K{=}128$: a mode-collapsed policy does not reach the rater-preferred regions, regardless of sample budget.

Therefore, We propose \method, Driving-Intent-Amplified reinforcement Learning, a two-stage training framework that first expands the sampling distribution of a continuous-action driving policy and then preserves this expansion during preference RL.
The first stage uses intent-conditioned CFG to decode the same scene into semantically distinct maneuver modes -- turning versus yielding, changing lanes versus keeping the lane, accelerating versus braking -- so the flow-matching action head produces alternatives beyond coordinate-level perturbations of one logged path.
With eight rule-derived intents (cruise, lane change L/R, turn L/R, U-turn, accelerate, decelerate), intent-conditioned best-of-$K$ already matches the human-driven demonstration at $K{\approx}8$, and pooling proposals across all eight intents reaches RFS $9.14$ at $K{=}128$.
The strongest prior end-to-end driving planner, RAP~\citep{feng2025rap}, caps at RFS $8.5$ even with best-of-$64$ over a learned rater; intent-conditioned sampling surpasses for the first time both this prior ceiling and the human-driven demonstration (RFS $8.13$).

The second stage of \method is needed because an expanded best-of-$N$ ceiling does not directly produce a deployable policy: oracle selection is unavailable at inference, and the policy still commits to one trajectory per frame.
Reinforcement fine-tuning is the natural tool for capturing the ceiling, but standard GRPO does not preserve the expansion.
Rollouts drawn from a single intent -- whether the GT, predicted, or top-rated -- re-collapse the rollout group around one mode, leaving group-relative advantages without preference contrast \citep{schulman2017proximal,shao2024deepseekmath}.
The bottleneck is the sampling distribution, not the policy update, as shown in Figure~\ref{fig:teaser}.

To preserve the diversity created by the first stage, \method uses \emph{multi-intent GRPO}.
In the first stage, we condition the flow-matching action head on the eight driving intents and train with classifier-free guidance dropout.
In the second stage, for each scene we sample a small number of trajectories under each intent, pool the eight intent-conditioned rollout sets into a single group, and compute group-relative advantages over the pooled set.
At deployment, a learned intent classifier selects the intent and the same intent-conditioned generator decodes the final trajectory, so the proposal mechanism used during RL remains active at inference.

On a deterministic 338/100 split of the RFS-labeled WOD-E2E validation pool, \method improves held-out RFS from $7.681$ to $8.211$ over its SFT initialization, while every single-intent variant peaks lower and declines by its final checkpoint.
Preference alignment for continuous-action policies trained from demonstrations is not the problem of imitating logged trajectories more accurately, nor of running a stronger optimizer; it is the problem of escaping mode collapse and preserving the escape through preference RL.

\textbf{Contributions.}
\begin{enumerate}[leftmargin=*,nosep]

  \item We identify single-demonstration SFT mode collapse as a key bottleneck for preference RL on continuous-action policies. On WOD-E2E, the best-of-$N$ RFS ceiling of competitive SFT baselines saturates below the logged human-driven demonstration even at $K{=}128$, showing that preference optimization is limited by what the policy can sample.

  \item We show that intent-conditioned classifier-free guidance breaks this sampling ceiling by expanding the flow-matching action head along semantically distinct maneuver modes. With eight rule-derived driving intents, intent-CFG sampling reaches best-of-$128$ RFS of $9.14$, surpassing for the first time both the prior best planner RAP (RFS $8.5$ with best-of-$64$) and the human-driven demonstration (RFS $8.13$).

  \item We introduce \method, Driving-Intent-Amplified Learning, a two-stage training framework that combines intent-CFG proposal expansion with multi-intent GRPO diversity preservation. By spanning all intent classes within every preference group, \method preserves the expanded sampling distribution during RL and improves held-out RFS from $7.681$ to $8.211$, while every single-intent variant peaks lower and declines by the end of training.
\end{enumerate}

%% file: sections/02_related_work.tex
\section{Related Work}
\label{sec:related_work}

\method connects four lines of work. End-to-end autonomous driving has moved from sensor-fusion imitation and scalable decoding toward planning-oriented, vectorized, sparse, generative, and diffusion-based planners \citep{hu2023planning,jiang2023vad,chen2024vadv2,sun2025sparsedrive,zheng2024genad,liao2025diffusiondrive,zou2025diffusiondrivev2,hu2022st,chitta2022transfuser,wu2022trajectory,shao2023safety,jia2023think}. These systems motivate planning as the central output, while WOD-E2E turns evaluation toward rater-preference scores rather than logged-trajectory imitation alone \citep{xu2025wod}.

Vision-language-action models couple perception, language-conditioned representations, and continuous control for robotics and driving \citep{zitkovich2023rt,kim2024openvla,black2024pi_0,zhou2025autovla}. In autonomous driving, related VLM, LLM, and VLA systems study interpretable driving, graph VQA, closed-loop language-conditioned driving, multimodal driving language models, knowledge-driven driving, counterfactual reasoning datasets, reasoning-to-planning RL, latent-action prediction, scene-adaptive experts, style-aware actions, world models, and VLA reasoning \citep{xu2024drivegpt4,sima2024drivelm,shao2024lmdrive,ma2024dolphins,mao2023gpt,wen2023dilu,wang2025omnidrive,zhou2026opendrivevla,li2026drive,zhang2025openread,rawal2026nord,xie2026latentvla,luo2026last,you2026samoe,gao2026stylevla,zhang2025reasoning,wang2026learning,li2025drivevla,ye2025vla}. Our work is not primarily a larger driving VLA; it isolates when preference optimization has useful maneuver-level proposals to rank.

Multimodal driving prediction has long represented future behavior through anchors, trajectory sets, latent modes, or intention queries \citep{chai2019multipath,phan2020covernet,salzmann2020trajectron++,shi2022motion}. We use intent for a different role, namely to enlarge the maneuver-level proposal support available to a continuous ego policy.

The optimization builds on PPO/RLHF-style preference learning and group-relative policy updates \citep{schulman2017proximal,ziegler2019fine,rafailov2023direct,shao2024deepseekmath}, as well as recent diffusion and flow policy optimization \citep{chi2025diffusion,lipman2022flow,ren2024diffusion,alles2025flowq,zhang2025reinflow,mcallister2025flow,black2023training,yang2024using,wang2022diffusion,huang2026tide}. Our focus is the interaction between intent-diverse proposals and \rfs-guided reinforcement learning for planning-oriented continuous-action \vla driving. A fuller discussion is provided in Appendix~\ref{app:related_work}.

%% file: sections/03_method.tex
\section{Method}
\label{sec:method}

\method is a two-stage training pipeline built on top of \backbone \citep{huang2026mindvla}. 
Given a driving scene $x$, the policy produces a future trajectory $\tau \in \mathbb{R}^{T \times d}$, and under \rfs supervision the logged trajectory is one scored candidate rather than an oracle target.
Stage~1 conditions the diffusion action head on a discrete driving intent with classifier-free guidance, expanding the sampling distribution along intent-level axes.
Stage~2 runs GRPO over an intent-balanced rollout group that spans the full intent set per scene, preserving intent diversity through preference optimization. 
At inference, a lightweight intent classifier predicts the intent from visual context, and the same intent-conditioned generator decodes the final trajectory.

\begin{figure*}[t]
  \centering
  \includegraphics[width=\linewidth]{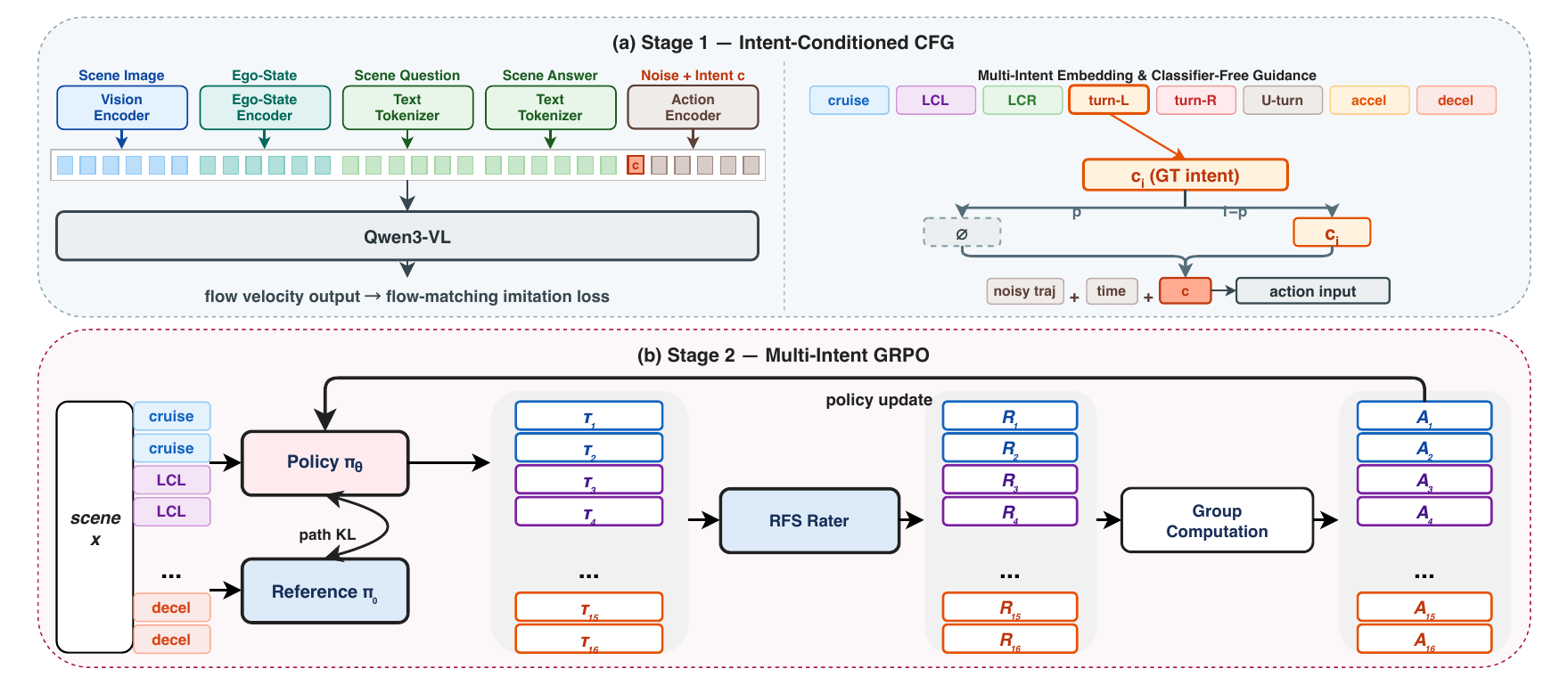}
  \caption{
    \textbf{Overview of \method.} 
    \textbf{(a) Stage 1 — CFG Imitation Training.} The diffusion action head is conditioned on a discrete intent $c_i$; CFG dropout ($p_{\mathrm{drop}}$) teaches the model both conditional and unconditional action distributions. \textbf{(b) Stage 2 — Multi-Intent GRPO.} Per scene, $K{=}16$ trajectories are sampled as $S{=}2$ rollouts $\times$ $|\mathcal{C}|{=}8$ intents, scored by the RFS rater, and used to update the policy via GRPO against the SFT reference $\pi_0$.
  }
  \label{fig:method}
\end{figure*}

\subsection{Intent-Conditioned CFG}
\label{subsec:intent_proposals}

We condition the flow-matching action head on a discrete driving intent $c$.
The intent set is deliberately small: cruise, lane change left/right, turn left/right, U-turn, accelerate, and decelerate.
During imitation training, these labels are inferred from trajectory geometry using displacement, heading change, lateral shift, and speed change.
An intent embedding is added to the noisy-action/time suffix before the language model predicts the flow velocity.
Classifier-free dropout also replaces some intent labels with an unconditional placeholder, so the same generator learns both conditional and unconditional action distributions.
At inference, a lightweight prefix classifier predicts the intent from the visual-state context and CFG combines conditional and unconditional velocity predictions.

This stage is not meant to encode a complete driving ontology.
Its purpose is to prevent single-trajectory SFT from collapsing distinct intent basins into one continuation.
For the same scene, changing $c$ should change the proposal at the level raters can judge, not merely add coordinate noise.

\subsection{Multi-Intent GRPO}
\label{subsec:rfs_grpo}

For each scene, preference optimization is applied to an intent-balanced proposal group that spans the entire intent set, not to multiple noise seeds drawn from a single intent.
We sample $S$ trajectories per intent, giving
\begin{equation}
  K = |\mathcal{C}| S
\end{equation}
proposals.
The current implementation uses $|\mathcal{C}|=8$ and $S=2$, so each scene contributes $K=16$ trajectories.
Each trajectory is generated by the stochastic SDE sampler of the flow-matching head, and its cumulative path log-probability is stored for policy-ratio replay.

The reward is the rater feedback score:
\begin{equation}
  R_i = \rfs(\tau_i; \mathcal{P}_x),
\end{equation}
where $\mathcal{P}_x$ denotes the rater trajectories and labels for scene $x$.
Auxiliary geometric costs can be included, but the main implementation keeps them at zero and uses the preference reward described below.
Advantages are normalized within the proposals of the same scene:
\begin{equation}
  A_i = \frac{R_i - \frac{1}{K}\sum_{j=1}^{K} R_j}
             {\operatorname{std}_{j=1}^{K}(R_j) + \epsilon}.
\end{equation}
This is the step that makes intent diversity useful for RL: if all proposals express the same intent, the group has little preference contrast; if intents expose different intents, RFS supplies a meaningful within-scene ranking.
Because the group always spans all eight intents rather than $K$ noise variants of one, every intent class receives a preference signal in the same update.

\subsection{Reward-Hacking-Aware RFS Reward}
\label{subsec:reward_hacking_reward}

The reported evaluation metric remains the standard WOD-E2E RFS score.
For training, however, using the evaluator unchanged creates two avoidable reward-hacking paths.
Standard RFS scores only the $3$s and $5$s anchors, so intermediate waypoints can drift while the two anchors stay inside the acceptable boxes.
It also applies a hard maximum over up to three rater trajectories, so a model can overfit to the easiest high-label rater geometry on the RL training split.

We therefore use a training-side RFS variant with the same rater preference inputs but different aggregation.
First, the rater maximum is replaced by label-softmax aggregation.
For rater label $y_p$, geometric decay $d_{p,a}$ at anchor $a$, and temperature $\tau$, the per-anchor score is
\begin{equation}
  \widetilde{R}_a
  =
  \sum_p
  \frac{\exp(\tau y_p)}{\sum_q \exp(\tau y_q)}
  y_p d_{p,a}.
\end{equation}
The weights depend only on fixed rater labels, not on model-controlled geometry, so the policy cannot manipulate the aggregation by moving toward one rater. We tune $\tau$ on the split-aware validation protocol and report the sensitivity in Section~\ref{subsec:reward_hacking_ablation}.
Second, the training anchors are densified from $\{3,5\}$ seconds to $\{1,2,3,4,5\}$ seconds.
The additional anchors penalize discontinuous or implausible intermediate motion while preserving the canonical $3$s and $5$s scoring points used by evaluation.

\subsection{Policy Update}
\label{subsec:policy_update}

Each sampled trajectory is replayed under the same intent condition used to generate it.
Let $\log p_{\theta_{\mathrm{old}}}(\tau_i \mid x,c_i)$ be the stored SDE path log-probability and $\log p_{\theta}(\tau_i \mid x,c_i)$ the replayed current log-probability.
The clipped \grpo objective is
\begin{align}
  \rho_i =
  \exp\!\left(
    \log p_{\theta}(\tau_i \mid x,c_i)
    -
    \log p_{\theta_{\mathrm{old}}}(\tau_i \mid x,c_i)
  \right), \\
  \mathcal{L}_{\mathrm{GRPO}}
  =
  -\frac{1}{K}\sum_{i=1}^{K}
  \min\!\left(
    \rho_i A_i,
    \operatorname{clip}(\rho_i, 1-\epsilon_{\ell}, 1+\epsilon_{h}) A_i
  \right).
\end{align}
We use symmetric clipping with $\epsilon_{\ell}=\epsilon_h=0.2$.
The update is regularized against the starting \sft policy $\theta_0$ using a reference path penalty,
\begin{equation}
  \mathcal{R}_{\mathrm{ref}}
  =
  \exp(\Delta_i) - \Delta_i - 1,
  \quad
  \Delta_i =
  \log p_{\theta_0}(\tau_i \mid x,c_i)
  -
  \log p_{\theta}(\tau_i \mid x,c_i).
\end{equation}
The final loss is $\mathcal{L}_{\mathrm{GRPO}} + \beta \mathcal{R}_{\mathrm{ref}}$ with $\beta=0.002$.
$\mathcal{R}_{\mathrm{ref}}$ is the standard $k_3$ KL estimator computed on the cumulative SDE path log-probability and acts as a vanilla reference-policy penalty.

%% file: sections/04_experiments.tex
\section{Experiments}
\label{sec:experiments}

\subsection{Experimental Setup}
\label{subsec:exp_setup}

We evaluate \method on \backbone under the WOD-E2E \rfs protocol.
The first-stage \sft model is trained on the Waymo training split.
Preference optimization uses the \rfs-labeled validation pool.
We split the 438 validation sequences by a deterministic hash of \texttt{sequence\_id}: 338 sequences are used for RL training and 100 sequences are held out for evaluation, with \texttt{split\_seed=43}.
All RL comparisons below use this split.

The main run starts from the \texttt{ckpt8k} intent-conditioned \sft checkpoint and applies multi-intent CFG \grpo with $C=8$ intents and $S=2$ samples per intent, giving $K=CS=16$ proposals per scene.
Training uses batch size $4$, constant learning rate $5\times 10^{-7}$, CPS SDE sampling with noise level $0.5$, PPO clipping $0.2$, and reference-path coefficient $\beta=0.002$ on the cumulative-log-probability $k_3$ KL estimator.
The training-side reward uses the reward-hacking-aware \rfs variant described in Section~\ref{subsec:reward_hacking_reward}; evaluation reports standard Waymo \rfs.
Held-out \rfs is the main selection metric; full-split \rfs and trust-region rate (TR) describe the selected checkpoint.

\subsection{Main Results}
\label{subsec:main_results}

\textbf{Baseline training protocol.} For Table~\ref{tab:main_results}, we retrain and adapt the official baseline implementations under a common Waymo-only task-training protocol.
The supervised fine-tuning stage is constructed exclusively from the Waymo training split through our Waymo dataloader, which provides camera observations, ego-motion context, navigation intent, and future trajectory supervision, and converts them into the dataset format required by each model.
The same Waymo \sft source is used for ReCogDrive~\citep{li2025recogdrive}, AutoVLA~\citep{zhou2025autovla}, Curious-VLA~\citep{chen2026devil}, and WAM-Flow~\citep{xu2025wam}; no non-Waymo driving dataset is used for task-specific supervised fine-tuning.
When RL is applied, each policy is initialized from its Waymo-\sft checkpoint and optimized on the Waymo \rfs subset with \grpo-style rollouts and the MindVLA/Waymo \rfs reward.
Thus, official baseline codebases and released weights serve only as architectural implementations or initializations, while all task-specific \sft data and RL reward data in our experiments come from Waymo.

\textbf{Claim.} Under this controlled Waymo-only task-training protocol, \method produces the strongest held-out improvement among the RL-trained systems in Table~\ref{tab:main_results}.
Each peak is compared with the step-0 evaluation from the same RL stage, so the reported gain isolates preference optimization rather than SFT training progress.
Starting from \backbone at $7.696$ held-out \rfs, \method reaches $8.211$ ($+0.515$), with full-split \rfs rising from $7.369$ to $8.631$.
The baselines improve less on the same held-out split: WAM-Flow gains $+0.087$, Curious-VLA $+0.146$, AutoVLA $+0.043$, and ReCogDrive $+0.315$.
\method also has the highest peak held-out score and the largest TR increase ($54.7\%$ to $68.0\%$), while ReCogDrive is the strongest baseline peak at $7.714$.
These comparisons are consistent with the paper's central mechanism: preference optimization is most effective when intent-conditioned sampling exposes maneuver-level alternatives that \rfs{} can rank, rather than only local coordinate variants of a logged path.

\begin{table}[h]
  \centering
  \small
  \caption{WOD-E2E \rfs on the 338/100 held-out split (\texttt{split\_seed=43}) after Waymo-only \sft and preference optimization. All models use the same task-specific Waymo \sft source. Within each block, the peak row is the RL checkpoint with the highest held-out \rfs and the init row is the step-0 evaluation from the same RL stage; $\Delta_{\mathrm{RL}}$ is the held-\rfs change from init; TR and Full \rfs are at the peak checkpoint.}
  \label{tab:main_results}
  \begin{tabular*}{\textwidth}{@{\extracolsep{\fill}}llccccc}
    \toprule
    Model & Action Representation & Stage & Held \rfs & $\Delta_{\mathrm{RL}}$ & TR & Full \rfs \\
    \midrule
    \multirow{2}{*}{WAM-Flow}    & \multirow{2}{*}{discrete flow}   & \sft init  & 5.547 & --     & 24.0\% & 5.757 \\
                                 &                                  & RL peak    & 5.634 & +0.087 & 19.0\% & 6.111 \\
    \addlinespace[0.2em]
    \multirow{2}{*}{Curious-VLA} & \multirow{2}{*}{action token}    & \sft init  & 5.808 & --     & 30.0\% & 5.827 \\
                                 &                                  & RL peak    & 5.954 & +0.146 & 31.0\% & 7.157 \\
    \addlinespace[0.2em]
    \multirow{2}{*}{AutoVLA}     & \multirow{2}{*}{action token}    & \sft init  & 6.744 & --     & 46.0\% & 6.809 \\
                                 &                                  & RL peak    & 6.787 & +0.043 & 47.0\% & 6.780 \\
    \addlinespace[0.2em]
    \multirow{2}{*}{ReCogDrive}  & \multirow{2}{*}{DiT diffusion}   & \sft init  & 7.399 & --     & 58.0\% & 7.244 \\
                                 &                                  & RL peak    & 7.714 & +0.315 & 65.4\% & 7.543 \\
    \midrule
    \multirow{2}{*}{DIAL}        & \multirow{2}{*}{continuous flow} & \sft init  & 7.696 & --     & 54.7\% & 7.369 \\
                                 &                                  & RL peak    & 8.211 & +0.515 & 68.0\% & 8.631 \\
    \bottomrule
  \end{tabular*}
\end{table}
\FloatBarrier

\subsection{Intent-CFG Proposal Ceiling}
\label{subsec:proposal_ceiling}

\textbf{Claim.} Intent-conditioned CFG expands the proposal support beyond ordinary \sft sampling, raising the best-of-$N$ \rfs ceiling above the logged human demonstration before any RL update.

\begin{figure}[t]
  \begin{minipage}[t]{0.49\linewidth}
    \centering
    \includegraphics[width=\linewidth]{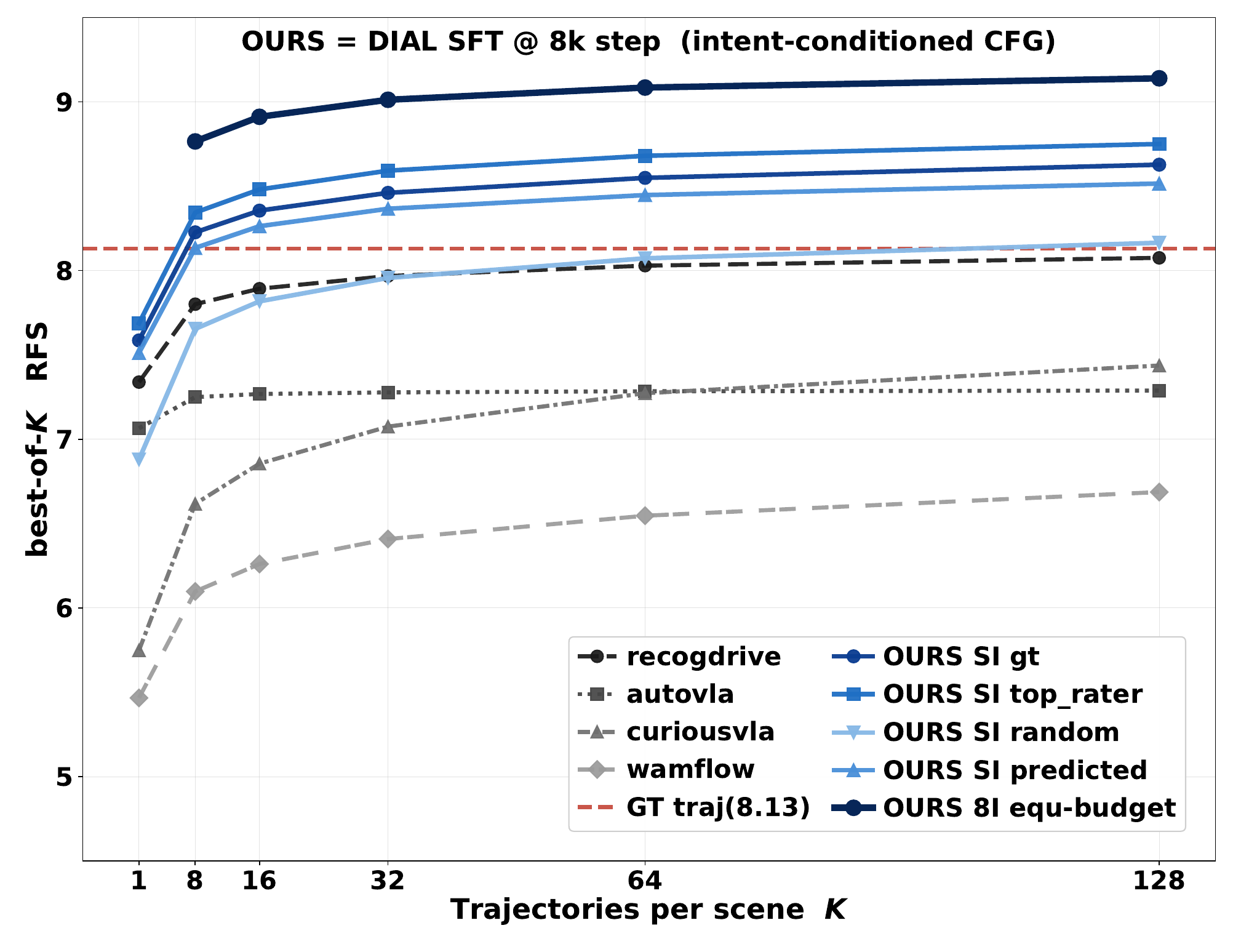}
    \caption{%
      \textbf{Pre-RL proposal ceiling.}
      Best-of-$K$ \rfs vs.\ budget $K$.
      Gray dashed: four SFT baselines all saturate below GT ($8.13$, red dashed) at $K{=}128$.
      Blue: intent-conditioned SFT under four strategies (\textit{gt}, \textit{top-rater}, \textit{predicted}, \textit{random}), all cross GT at $K{\approx}8$.
      Navy: 8-intent equal-budget pooling reaches $9.14$ at $K{=}128$.
    }
    \label{fig:bon_intent_strategies}
  \end{minipage}\hfill
  \begin{minipage}[t]{0.49\linewidth}
    \centering
    \includegraphics[width=\linewidth]{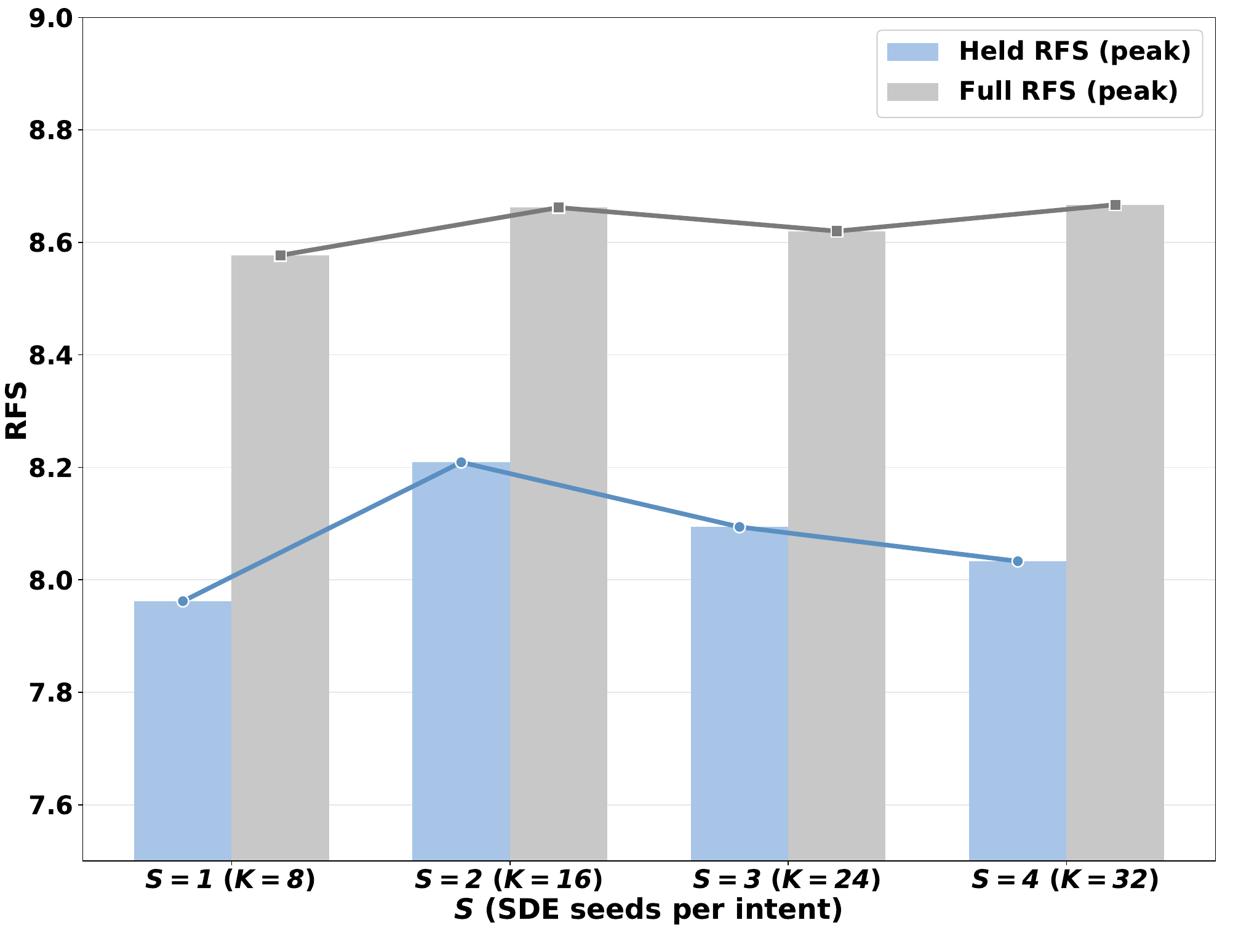}
    \caption{%
      \textbf{Samples per intent at $C{=}8$.}
      Sweeping $S\in\{1,2,3,4\}$ ($K\in\{8,16,24,32\}$):
      held-out \rfs (light blue) is non-monotone with $S{=}2$ as the peak-height sweet spot ($8.21$);
      full-split \rfs (light gray) is nearly flat across $S$.
    }
    \label{fig:samples_per_intent}
  \end{minipage}
\end{figure}

On the same \rfs-labeled scenes, we compare three proposal supports: the logged trajectory, best-of-$N$ from the ordinary \sft policy (four competitive VLA baselines: WAM-Flow, Curious-VLA, AutoVLA, ReCogDrive), and best-of-$N$ from the intent-conditioned \sft policy under four per-sample intent strategies.
Figure~\ref{fig:bon_intent_strategies} plots the expected best-of-$K$ \rfs as $K$ grows from $1$ to $128$.
The logged trajectory scores $8.13$ \rfs (red dashed reference).
All four baseline VLAs saturate \emph{below} the GT line at $K=128$, confirming that ordinary stochastic sampling from the \sft policy cannot recover trajectories the human rater would prefer over the log.
The intent-conditioned policy with any single-intent strategy (gt, classifier-predicted, top-rater, or random) crosses the GT line already at $K\approx8$, because intent conditioning guides proposals into distinct semantic basins rather than perturbing a single maneuver mode.
Pooling proposals across all eight intent classes with equal per-intent budget (right-most curve, $K = 8 \times n_\text{per-intent}$) extends the ceiling to $9.14$ at $K=128$, an improvement of $+1.07$ over the best baseline.

\FloatBarrier

\subsection{Multi-Intent vs Single-Intent}
\label{subsec:multi_vs_single}

\textbf{Claim.} Spanning the full intent set within each preference group is necessary; any single-intent recipe -- regardless of how the conditioning intent is selected -- either collapses or plateaus before \method's peak.

We hold all other settings fixed at the main \method recipe and vary only how each rollout's conditioning intent is chosen, while keeping the per-scene budget at $K=16$.
Multi-intent \grpo (the main run) draws $K=16$ rollouts per scene by spanning all $8$ intent classes with $S=2$ per intent.
The four single-intent baselines use $C=1$ and $S=16$ noise seeds, so the total budget remains $K=16$.
The single-intent variant is then defined by how the one intent is picked: \textit{gt} from the geometric intent of the logged trajectory; \textit{predicted} from the deployment intent classifier head; \textit{top-rater} from the highest-rated rater trajectory's geometry (a label-leakage upper bound); and \textit{random} from a uniform draw at every batch.

Table~\ref{tab:multi_vs_single} compares held-out \rfs peaks across group compositions at the same total budget $K=16$.
\method reaches a peak of $8.211$ ($+0.515$ vs \sft init).
None of the four single-intent baselines reaches $8.00$, and all ultimately decline.
\textit{Random}, the strongest single-intent baseline, peaks at $7.992$ but falls $0.22$ short of \method.
\textit{Predicted} reaches $7.864$ with a delayed collapse; its training-split \rfs continues rising to $8.156$ while held-out \rfs falls, the standard signature of reward hacking.
\textit{Gt} and \textit{top-rater} collapse more sharply.
At \method's peak checkpoint, all single-intent variants lie $0.67$--$1.98$ \rfs points below.
Spanning the full intent set is therefore necessary both for reaching a competitive peak and for maintaining preference contrast throughout training.
Training dynamics for all variants are visualized in Figure~\ref{fig:multi_intent_train_curves} (Appendix~\ref{app:training_curves}).

\begin{table}[!htbp]
  \centering
  \small
  \caption{Multi-intent group vs single-intent baselines, all sharing the main \method recipe and per-scene budget $K=16$. TR and Full \rfs are reported at each variant's held-peak checkpoint.}
  \label{tab:multi_vs_single}
  \begin{tabular*}{\textwidth}{@{\extracolsep{\fill}}lcccccc}
    \toprule
    Group composition & $C$ & $S$ & $K$ & Held peak & TR & Full \rfs \\
    \midrule
    \textit{gt} (single, geometric)            & 1 & 16 & 16 & 7.783 & 65.0\% & 7.733 \\
    \textit{predicted} (single, classifier)    & 1 & 16 & 16 & 7.864 & 57.0\% & 8.331 \\
    \textit{top-rater} (single, leakage)       & 1 & 16 & 16 & 7.728 & 61.0\% & 7.545 \\
    \textit{random} (single, uniform)          & 1 & 16 & 16 & 7.992 & 64.0\% & 8.035 \\
    \method (multi-intent, main)               & 8 & 2  & 16 & \textbf{8.211} & \textbf{68.0\%} & \textbf{8.631} \\
    \bottomrule
  \end{tabular*}
\end{table}
\FloatBarrier

\subsection{Diversity Preservation Analysis}
\label{subsec:diversity}

\textbf{Claim.} Multi-intent CFG \grpo prevents proposal diversity from collapsing during preference optimization, preserving the expanded sampling distribution established in Stage~1.

We support the claim with diversity metrics computed by inference on already-trained checkpoints.
No additional RL training is required.

\textbf{Inter-intent proposal distance (D1) and \rfs spread (D2).}
For each scene we decode under all $8$ intents (with CFG disabled to obtain pure intent-conditional samples), compute D1 as the mean pairwise ADE across the $\binom{8}{2}=28$ trajectory pairs, and D2 as the \rfs standard deviation across the $8$ trajectories.
All RL variants are evaluated at the same checkpoint (iter 4800) for a fair comparison.
Table~\ref{tab:diversity} shows that \method retains the highest D2 ($0.75$), meaning different intents still produce quality-differentiated trajectories after preference optimization.
Single-intent variants show two failure modes: \textit{random} collapses spatially (D1 $= 2.40$~m, lowest), producing near-identical trajectories regardless of the intent conditioning, while \textit{top-rater} shows spatial scatter (D1 $= 7.08$~m, highest) but poor quality differentiation (low D2).
The diversity dividend \emph{gap} $=$ D3@16 $-$ D3@1 measures how much additional \rfs is recoverable by selecting among $16$ intent-conditioned proposals.
\method preserves a gap of $+2.04$ (close to the SFT initialization's $+2.23$), while all single-intent variants lose between $0.19$ and $0.57$ gap versus \method.
Note that \method's D3@16 ($6.540$) is slightly below the SFT initialization's ($6.617$): RL concentrates probability toward better-scoring deployment modes at the cost of a small reduction in the best-of-$16$ diversity ceiling -- the expected exploitation--diversity trade-off, which does not affect greedy deployment (held-out \rfs improves from $7.696$ to $8.211$).
Crucially, the gap ordering matches the held-out \rfs ranking exactly: the more diversity is preserved during RL, the higher the final \rfs.

\begin{table}[!htbp]
  \centering
  \small
  \caption{Diversity metrics at iter 4800 for a fair cross-variant comparison. D1: mean pairwise ADE across 8 intent-conditional trajectories per scene. D2: per-scene \rfs std across the same 8 trajectories. D3@1/D3@16: best-of-1 and best-of-16 held-out \rfs. Gap $=$ D3@16 $-$ D3@1: diversity dividend. The SFT init row uses the \texttt{ckpt8k} checkpoint before any RL.}
  \label{tab:diversity}
  \begin{tabular*}{\textwidth}{@{\extracolsep{\fill}}lcccccc}
    \toprule
    Sampling & D1 (m) & D2 & D3@1 & D3@16 & Gap \\
    \midrule
    SFT init (no RL)       & 6.43 & 0.52 & 4.387 & 6.617 & +2.23 \\
    \midrule
    \method (multi-intent) & 4.17 & \textbf{0.75} & \textbf{4.500} & \textbf{6.540} & \textbf{+2.04} \\
    Single: random         & 2.40 & 0.43 & 4.381 & 6.231 & +1.85 \\
    Single: predicted      & 4.82 & 0.64 & 4.372 & 6.172 & +1.80 \\
    Single: top-rater      & 7.08 & 0.65 & 4.240 & 6.020 & +1.78 \\
    Single: GT intent      & 6.02 & 0.59 & 4.229 & 5.889 & +1.66 \\
    \bottomrule
  \end{tabular*}
\end{table}
\FloatBarrier

\subsection{Ablation Studies}
\label{subsec:ablation}

\subsubsection{Samples per Intent}
\label{subsec:samples_per_intent}

\textbf{Claim.} Holding the intent set at $C=8$ fixed, the number of samples per intent $S$ controls a trade-off between within-intent advantage variance and reward-hacking onset.

Figure~\ref{fig:samples_per_intent} sweeps $S\in\{1,2,3,4\}$ at fixed $C=8$, giving total group sizes $K\in\{8,16,24,32\}$.
$S=1$ ($K=8$) supplies one rollout per intent and gives the cheapest group; the held-out peak is $7.962$ and the run remains stable throughout training.
$S=2$ ($K=16$, the main configuration) reaches the highest peak $8.211$.
$S=3$ ($K=24$) reaches a peak of $8.094$ and $S=4$ ($K=32$) peaks at $8.033$, confirming that higher $S$ shifts the peak earlier and slightly reduces peak height.
Across all runs, $S=2$ remains the peak-height sweet spot; $S=1$ remains the deployment-stability operating point.

\subsubsection{Reward Shaping}
\label{subsec:reward_hacking_ablation}

Standard Waymo \rfs uses a hard maximum over rater trajectories and scores only the $3$s and $5$s anchors, creating avoidable reward-hacking paths during RL.
We replace the maximum with a label-softmax aggregation (temperature $\tau$) and densify the scored anchors to $\{1,2,3,4,5\}$s; Table~\ref{tab:reward_hacking_ablation} ablates these choices.
The peak-height ordering is non-monotone in $\tau$: $\tau=0.3$ gives the highest held-out peak ($8.211$), while row~D ($\tau=1.0$ + dense anchors) achieves the strongest full-split score ($8.728$) with substantially better end-of-training stability.
We report $\tau=0.3$ as the main configuration for peak performance; row~D is the deployment-friendly alternative.

\begin{table}[!htbp]
  \centering
  \small
  \caption{Training-side reward ablation on the 338/100 held-out split (§\ref{subsec:exp_setup}). All variants share the multi-intent main recipe ($C=8$, $S=2$, $K=16$). TR and \textit{Full peak} are reported at the held-peak checkpoint.}
  \label{tab:reward_hacking_ablation}
  \begin{tabular*}{\textwidth}{@{\extracolsep{\fill}}lcccccc}
    \toprule
    Reward variant & Aggr & Anchor & $\tau$ & Held peak & TR & Full peak \\
    \midrule
    A. Vanilla                              & max     & sparse & --   & 7.990 & 68.0\% & 8.328 \\
    B. Dense anchors only                   & max     & dense  & --   & 7.965 & 67.0\% & 8.430 \\
    C. Softmax only                         & softmax & sparse & 1.0  & 8.147 & 72.0\% & 8.485 \\
    D. Softmax + dense (deployment-friendly) & softmax & dense & 1.0  & 8.130 & 62.0\% & \textbf{8.728} \\
    Softmax + dense ($\tau{=}0.5$)          & softmax & dense  & 0.5  & 8.097 & 62.0\% & 8.634 \\
    \method (main)                          & softmax & dense  & 0.3  & \textbf{8.211} & \textbf{68.0\%} & 8.631 \\
    Mean ($\tau{\to}0$) + dense             & mean    & dense  & --   & 7.834 & 60.0\% & 8.098 \\
    \bottomrule
  \end{tabular*}
\end{table}
\FloatBarrier

%% file: sections/05_conclusion.tex
\section{Conclusion}
\label{sec:conclusion}

We identified single-demonstration \sft mode collapse as the primary bottleneck for preference RL on continuous-action driving policies: when sampled trajectories cluster around one maneuver mode, the group-relative reward signal carries no maneuver-level contrast, and RL cannot improve beyond what the sampling distribution already contains.
\method addresses this with a two-stage training pipeline.
Stage~1 conditions the diffusion action head on discrete driving intents with classifier-free guidance, expanding the sampling distribution into semantically distinct maneuver basins before any RL update.
Stage~2 runs multi-intent \grpo over an intent-balanced proposal group per scene, preserving the expanded distribution through preference optimization; any single-intent rollout recipe re-collapses the group and degrades by training end.
Together, the two stages lift the best-of-$N$ \rfs ceiling to $9.14$ at $K{=}128$ -- surpassing both the strongest prior planner and the human-driven demonstration -- and improve held-out single-trajectory \rfs from $7.696$ to $8.211$, with no single-intent variant reaching $8.00$.

\paragraph{Limitations.}
Several aspects of the current implementation bound the scope of the conclusions.
First, the eight driving intents are derived from trajectory geometry by hand-coded rules; they capture common maneuver categories but may not cover uncommon long-tail behaviors, and the labeling heuristic can misclassify ambiguous trajectories.
Second, RL training uses the $438$-sequence \rfs-labeled validation pool, which is small relative to the full Waymo training split; the reward signal may not fully represent the distribution of challenging or rare scenarios.

%% file: sections/A_related_work.tex
\section{Extended Related Work}
\label{app:related_work}

\paragraph{Vision-language-action policies.}
Vision-language-action models adapt large multimodal representations to action generation. RT-2 studies how web-scale vision-language pretraining can transfer to robot control \citep{zitkovich2023rt}; OpenVLA provides an open-source VLA model for robotic manipulation \citep{kim2024openvla}; and $\pi_0$ uses a flow-based action model for general robot control \citep{black2024pi_0}. In autonomous driving, WOD-E2E defines a vision-based end-to-end benchmark with challenging long-tail scenarios and rater-feedback evaluation \citep{xu2025wod}, while AutoVLA studies a VLA driving model with adaptive reasoning and reinforcement fine-tuning \citep{zhou2025autovla}. Our work is not primarily a larger VLA architecture. It isolates a proposal-support bottleneck: a continuous-action VLA can only benefit from preference RL if it first proposes maneuver alternatives that the preference signal can rank.

\paragraph{Intent and multimodal trajectory proposals.}
Driving behavior is intrinsically multimodal. MultiPath predicts a distribution over anchor trajectory hypotheses \citep{chai2019multipath}, CoverNet casts behavior prediction as classification over a representative trajectory set \citep{phan2020covernet}, Trajectron++ uses a graph-structured generative model for dynamically feasible multi-agent forecasting \citep{salzmann2020trajectron++}, and Motion Transformer combines global intention localization with local trajectory refinement \citep{shi2022motion}. These methods show that future motion is better represented as a structured set of alternatives than as a single regression target. \method transfers that lesson to ego-policy optimization: the intent label is not the final prediction target, but a control variable that forces the proposal generator to expose distinct maneuvers before \rfs-guided RL.

\paragraph{Generative continuous-action policies.}
Diffusion and flow models are attractive action heads because they can represent complex continuous trajectory distributions. Diffusion Policy formulates visuomotor control as conditional action diffusion \citep{chi2025diffusion}, and flow matching trains continuous normalizing flows by regressing velocity fields along probability paths \citep{lipman2022flow}. Recent policy-optimization methods adapt these generators to reward optimization, including DPPO for diffusion policies \citep{ren2024diffusion}, FlowQ for offline RL with energy-guided flow policies \citep{alles2025flowq}, ReinFlow for online fine-tuning of flow matching policies \citep{zhang2025reinflow}, and flow matching policy gradients \citep{mcallister2025flow}. Our setting is complementary: we retain a flow-style continuous generator, but structure its sampling group by driving intent so that the reward sees maneuver-level variation rather than only coordinate-level noise.

\paragraph{Preference optimization and group-relative updates.}
Preference optimization commonly regularizes a learned policy against a reference policy while increasing the likelihood of preferred outputs. PPO supplies the clipped policy-gradient backbone \citep{schulman2017proximal}, RLHF-style fine-tuning applies learned human preference rewards to language models \citep{ziegler2019fine}, DPO removes explicit reward-model RL by optimizing pairwise preferences directly \citep{rafailov2023direct}, and GRPO uses group-relative advantages to reduce the need for a learned value function \citep{shao2024deepseekmath}. WOD-E2E's \rfs provides a planning-facing preference signal for trajectories rather than text \citep{xu2025wod}. \method therefore keeps the group-relative update, but defines the group as intent-diverse continuous proposals from the same scene and regularizes the update on the replayed continuous generation path.

%% file: sections/B_training_curves.tex
\section{Training Dynamics}
\label{app:training_curves}

\begin{figure}[h]
  \centering
  \includegraphics[width=\linewidth]{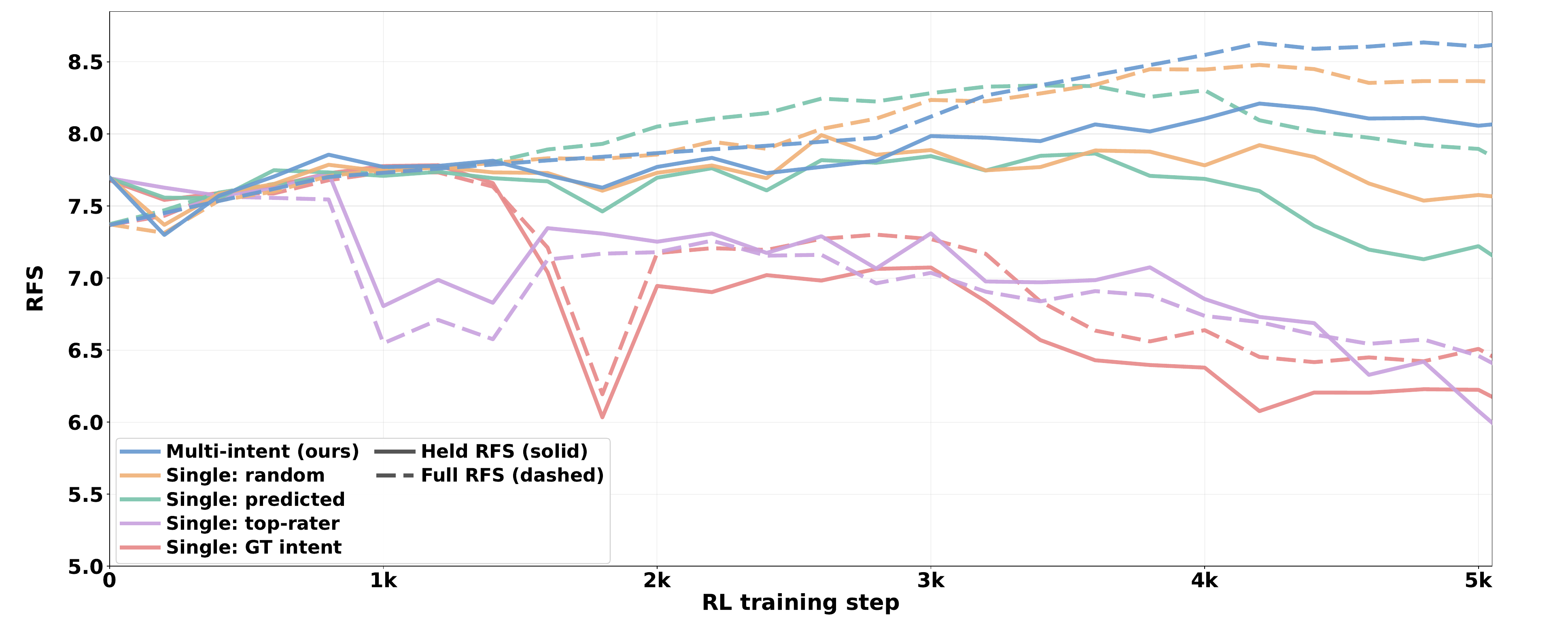}
  \caption{
    \textbf{Held-out \rfs throughout RL training.}
    \method (multi-intent, $C=8$, $S=2$) peaks highest and declines least among all variants.
    Single-intent baselines ($C=1$, $S=16$) all peak lower and collapse more sharply.
    \textit{Predicted} (dashed) shows the reward-hacking signature: training-split \rfs continues rising while held-out \rfs falls after the peak.
    All variants share the same per-scene budget $K=16$ and are evaluated on the 338/100 held-out split.
  }
  \label{fig:multi_intent_train_curves}
\end{figure}

Figure~\ref{fig:multi_intent_train_curves} plots held-out \rfs throughout RL training for \method and the four single-intent baselines from Section~\ref{subsec:multi_vs_single}, all sharing the same per-scene budget $K=16$.

Three patterns are visible.
First, \method (multi-intent) rises to its peak held-out \rfs of $8.211$ and subsequently declines only modestly, maintaining a substantially higher level than all single-intent variants throughout.
The relative stability reflects that intent-diverse rollouts supply meaningful within-group preference contrast at every iteration: the group always spans semantically distinct maneuvers, so the reward signal never saturates.

Second, all four single-intent variants peak lower and exhibit sharper post-peak declines.
\textit{Random} reaches the highest single-intent peak ($7.992$) but falls well below \method by training end.
\textit{Gt} and \textit{top-rater} collapse earliest, consistent with the observation that a fixed intent quickly exhausts preference contrast once the policy learns to score well on that intent.

Third, \textit{predicted} displays the canonical reward-hacking signature: training-split \rfs continues rising after the held-out peak while held-out \rfs falls, confirming that optimizing a single intent aligned with the classifier's prediction overfits to the RL training split rather than learning transferable preferences.
This divergence between training-split and held-out \rfs is absent in \method, where spanning all intents provides a structural barrier against distribution-specific overfitting.

